\documentclass[english]{article}
\usepackage[T1]{fontenc}
\usepackage[latin9]{inputenc}

\usepackage{babel}
\usepackage{array}
\usepackage{multirow}
\usepackage{graphicx}
\usepackage{hyperref}
\usepackage{wrapfig}
\usepackage[table]{xcolor}
\usepackage{booktabs}
\usepackage{amsfonts}
\usepackage{nicefrac}
\usepackage{microtype}
\usepackage{authblk}
\usepackage{float}
\makeatother

\begin{document}

\title{Medical Image Segmentation Based on Multi-Modal Convolutional Neural Network:\\
Study on Image Fusion Schemes}

\author[1]{Zhe Guo*}
\author[2]{Xiang Li*}
\author[3]{Heng Huang}
\author[1]{Ning Guo}
\author[1]{Quanzheng Li}
\affil[1]{Massachusetts General Hospital}
\affil[2]{Beijing Institute of Technology}
\affil[3]{University of Pittsburgh  *Joint first authors}

\maketitle

\begin{abstract}
Image analysis using more than one modality (i.e. multi-modal) has been increasingly applied in the field of biomedical imaging. One of the challenges in performing the multi-modal analysis is that there exist multiple schemes for fusing the information from different modalities, where such schemes are application-dependent and lack a unified framework to guide their designs. In this work we firstly propose a conceptual architecture for the image fusion schemes in supervised biomedical image analysis: fusing at the feature level, fusing at the classifier level, and fusing at the decision-making level. Further, motivated by the recent success in applying deep learning for natural image analysis, we implement the three image fusion schemes above based on the Convolutional Neural Network (CNN) with varied structures, and combined into a single framework. The proposed image segmentation framework is capable of analyzing the multi-modality images using different fusing schemes simultaneously. The framework is applied to detect the presence of soft tissue sarcoma from the combination of Magnetic Resonance Imaging (MRI), Computed Tomography (CT) and Positron Emission Tomography (PET) images. It is found from the results that while all the fusion schemes outperform the single-modality schemes, fusing at the feature level can generally achieve the best performance in terms of both accuracy and computational cost, but also suffers from the decreased robustness in the presence of large errors in any image modalities.
\end{abstract}
 
\section{Introduction}

In the field of biomedical imaging and computer-aided diagnosis, image acquisition and analysis with more than one modality (i.e. multi-modal) has been a research focus for years as different modalities encompass abundant information which can be complementary to each other. As described in one of the multi-modal imaging project for brain tumor segmentation \cite{RN908}, each of the modality can reveal a unique type of biological information for the tumor-induced tissue changes and poses "somewhat different information processing tasks". Certain multi-modal imaging techniques, such as the combination of Positron Emission Tomography (PET) and Computed Tomography (CT), have become the standard for clinical practice. Other techniques such as the simultaneous recording of functional Magnetic Resonance Imaging (fMRI) and electroencephalography (EEG) \cite{RN916} have been widely applied for neuroscience studies.

With the growing amount of multi-modal image data and the methods developed for analyzing them, one major yet largely under-discussed problem is that most of the reported methods are conceptually similar, but only dealing with a specific image processing problem and/or a specific scenario. As described in the review work of \cite{RN907}, any classical image fusion method are composed of "registration and fusion of features from the registered images". Motivated by the need for a unified framework to guide the methodology design for multi-modal image processing, we advance one step further from the abstraction of image fusion methods in \cite{RN907} and propose a conceptual architecture for image fusion that can cover most of the supervised multi-modal biomedical image analysis methods. The architecture consists of three image fusion schemes based on the main stages of any learning models: fusing at the feature level, fusing at the classifier level, and the fusing at the decision-making level. 

Further, we propose a novel multi-modal image segmentation framework based on Convolutional Neural Network (CNN). As a typical CNN consists of stages including feature extraction (in its convolutional layers), classification (in its fully connected layers) and decision making (in its output), we implement the three image fusion schemes into three corresponding image segmentation networks with varied structures. The proposed framework, which is the combination of the three segmentation networks, is applied on the multi-modal soft tissue sarcoma imaging dataset \cite{RN901}. Preliminary testing results show that all the image fusion schemes can outperform the single-modality schemes, which validates the effectiveness of multi-modal image segmentation method. By examining the performance difference across different fusion schemes and the cause thereof, we then provide several insights into the characteristics of the multi-modal feature learning and the impact of errors on the learning process. 
 
\section{Materials and Methods}

\subsection{Conceptual design for image fusion schemes}

As any supervised learning-based method conceptually consists of three steps: feature extraction, classification and decision making, we propose three schemes for fusing information from different image modalities as below: 
\begin{itemize}
\item Fusing at feature level: multi-modality images are used together to learn a unified image feature set. The feature set contains the intrinsic multi-modal representation of the data, which is used to train a traditional classifier. 
\item Fusing at the classifier level: images from each modality are used to learn a separate feature set of its own. These feature sets are then concatenated into a single feature set and used to train a multi-modal classifier.
\item Fusing at the decision level: images of each modality are used fully independently to train a single-modality classifier and learn the corresponding feature set. The final decision of this scheme is the fuse of the output from all the classifiers.
\end{itemize}

As these three schemes are the conceptual abstraction of the multi-modal image analysis, most of the current literature reports can be grouped accordingly. To name a few, works in \cite{RN763} (co-analysis of fMRI and EEG using CCA), \cite{RN906} (co-analysis of MRI and PET using PLSR) and \cite{RN909} (co-learning features through pulse-coupled NN) perform the feature-level fusion of the images. Works in \cite{RN904} (using contourlet), \cite{RN905} (using wavelet), \cite{RN910} (using wavelet) and \cite{RN896} (using features learned by LDS) perform the classifier-level fusion. The multi-resolution network proposed in \cite{RN839} performs fusion of images from multiple resolutions (rather than modalities) to utilize both local and global contrast within the images and fuse bottom-up saliency and high-level cues for better eye fixations prediction. Several works in image segmentation such as \cite{RN912} (fusing the results from different atlas by majority voting) and \cite{RN913} (fusing the SVM results from different modalities by majority voting), as well as the BRATS framework \cite{RN908} (using majority vote for fusing results from different algorithms, rather than modalities) belong to the decision-level fusion. To the best of our knowledge, no works have been done to empirically investigate the three fusion schemes simultaneously on the same dataset, as performed in our work.

\subsection{Data acquisition and preprocessing}

In this work, we use the Soft-tissue-Sarcoma dataset \cite{RN901} from the Cancer Imaging Archive (TCIA) \cite{RN902} for model development and validation. The dataset contains FDG-PET/CT as well as anatomical MR imaging (T1-weighted and T2-weighted fat-saturated) data from 50 patients with histologically proven soft-tissue sarcomas of the extremities. The FDG-PET transaxial resolution was 3.27mm and tangential resolution was 5.47mm. The median in-plane resolution for MR T1 imaging was 0.74mm (5.5mm thickness) and that of T2 was 0.63mm (5.0mm thickness). The tumor boundary was manually annotated by physicians. Annotations on the other modalities (PET/CT and T1) were then obtained using rigid registration. Since the PET/CT images have a much larger fields of view (FOV), they were truncated to the same regions with MR images. PET images were linearly interpolated to the same resolution with other modalities. The input for the network consists of images from four modalities (PET, CT, T1 and T2 MR), and the size of four-modality images is consistent for each individual subject but may vary among the cohort. A sample input images set is illustrated in Fig. 1. Image patches of size 28*28 are then extracted from the multi-modal images, respectively. Patches are labeled as "positive" if its center pixel is within the annotation (i.e. tumor-positive) region, and labeled as "negative" otherwise. Totally around 10 million patches are extracted from images of all modalities.
 
 \begin{figure}[H]
	\begin{centering}
  		\includegraphics[width=0.9\textwidth]{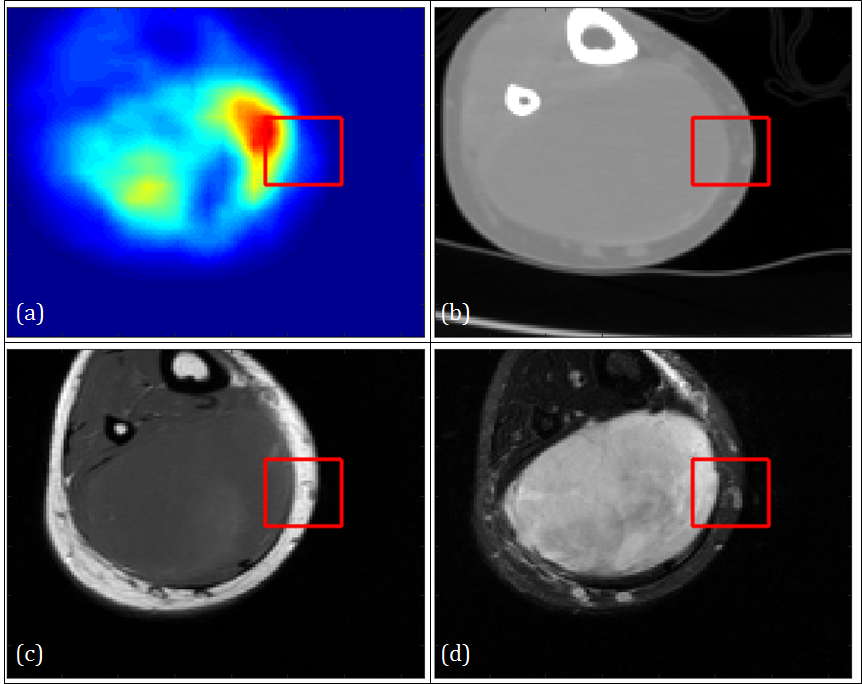}
  		\caption{Multi-modal images on the same position from a randomly-selected subject. (a): PET; (b): CT; (c): T1; (d): T2. The image size of this subject is 135$×$145. Red bounding box illustrates the size of patches (28$×$28) used as the input for CNN.}
  	\end{centering}
\end{figure}

\subsection{Multi-modal image classification using CNN}
 \begin{figure}[H]
	\begin{centering}
  		\includegraphics[width=1.0\textwidth]{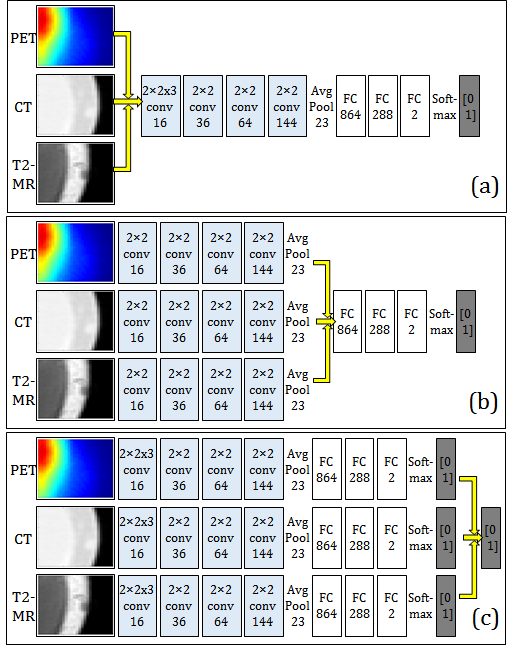}
  		\caption{Illustration of the structure for (a) Type-I fusion networks, (b) Type-II fusion network and (c) Type-III fusion network. The yellow arrows indicate the fusion location.}
  	\end{centering}
\end{figure} 

In this work we implement and compare the three image fusion schemes in three different patch-based CNNs with corresponding variations in their network structures as illustrated in Fig. 2: (a) Type-I fusion network implementing the feature-level fusing; (b) Type-II fusion network implementing the classifier-level fusing; and (c) Type-III fusion network implementing the decision-level fusing. In the figure we use the combination of PET, CT and T2-MR as an example for the fusion schemes, while for the experiment we also implement the networks using PET, CT and T1-MR, as well as the four modalities together as inputs. All the networks use the same set of image patches obtained in 2.2 as input. The network outputs, which are the labels of the given patches, are transformed into the labelmaps by assigning the corresponding label to the pixel in the patch center. We also train the single-modality CNNs for performance comparison.  

In Type-I fusion network, patches from different modalities are transformed into a 3-D tensor (28$×$28$×$k, where k is the number of modalities) and convoluted by a 2$×$2$×$k kernel as shown in Fig. 2(a). The 3-D kernel is then convoluted by a 2$×$2 kernel to fuse the high-dimensional features to the projected 2-D space, thus performs the feature-level fusion. In Type-II fusion network, the features are learned separately through each modality's own convolutional layers. The outputs of the last convolutional layer from each modality, which are considered the high-level representation of the corresponding images, are used to train a single fully connected network (i.e. classifier) as in Fig. 2(b). In Type-III fusion network, for each modality a typical single-modality 2-D CNN is trained. The prediction results (labelmaps) of these networks are then fused together based on majority vote (Fig. 3(c)) of the label on each pixel, to obtain the final labelmap. 

The image patch set derived from 50 subjects are divided into the training and testing sets for 10-folds cross-validation. In each run of the cross-validation, images from 45 patients are selected for training and the remaining 5 patients are used for testing. In each run, balanced numbers of "positive" and "negative" samples are used as the training input, while all the patches will be fed into the network during the testing. The statistics of the model performances are then reported based on the testing results.

\section{Experimental Results}

\subsection{Model performances comparison}

One sample result obtained by Type-I fusion network trained and tested on the combination of PET, CT and T2 images is visualized in Fig. 3. It can be observed that the CNN trained from the data can accurately predict the shape of the tumor region. Based on the labelmaps, the performance of each multi-modality and single-modality network is evaluated by their pixel-wise accuracy, as shown in Fig. 4. 

\begin{figure}[H]
	\begin{centering}
		\includegraphics[width=1.0\textwidth]{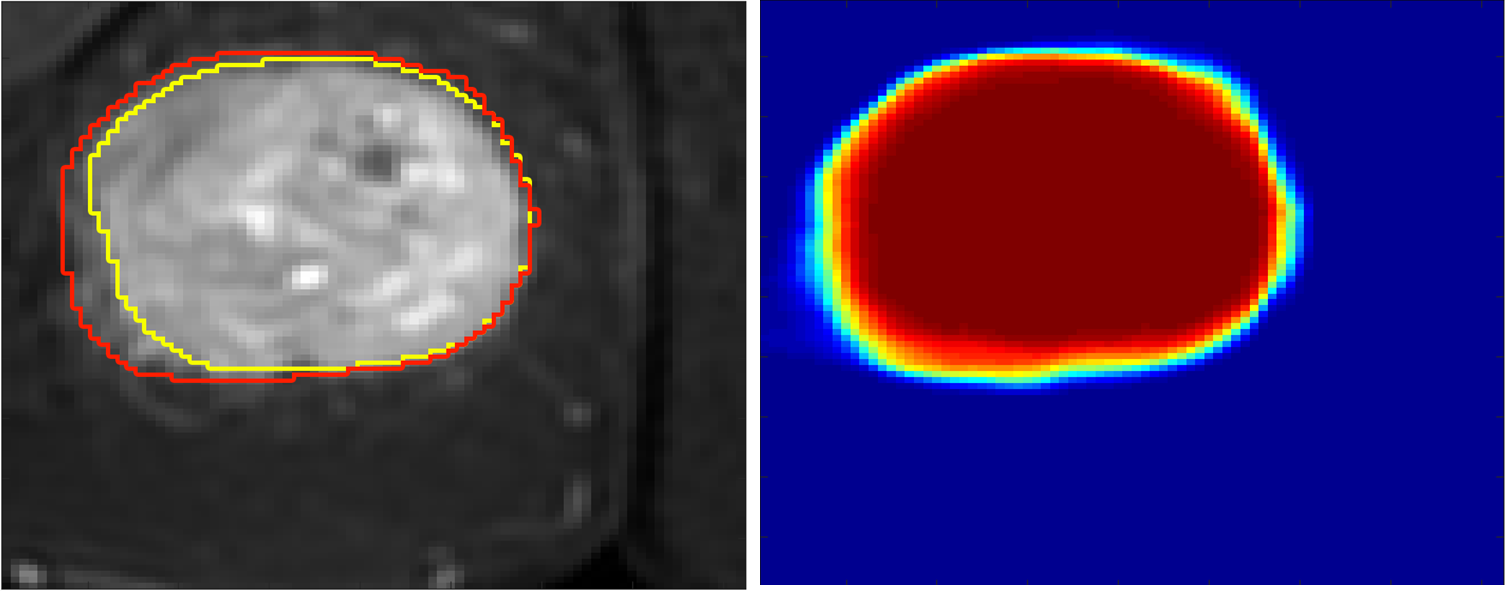}
		\caption{Left: Contour line of the ground truth annotation (yellow line) and labelmap (red line) overlaid on the T2-weighted MR image from one randomly-selected subject. Right: Heatmap of the network output on the same image, color-coded by the prediction probabilities of each pixel. Pixel with probability>0.5 will be predicted as "positive" and vice versa. }
	\end{centering}
\end{figure}

\begin{figure}[H]
	\begin{centering}
		\includegraphics[width=1.0\textwidth]{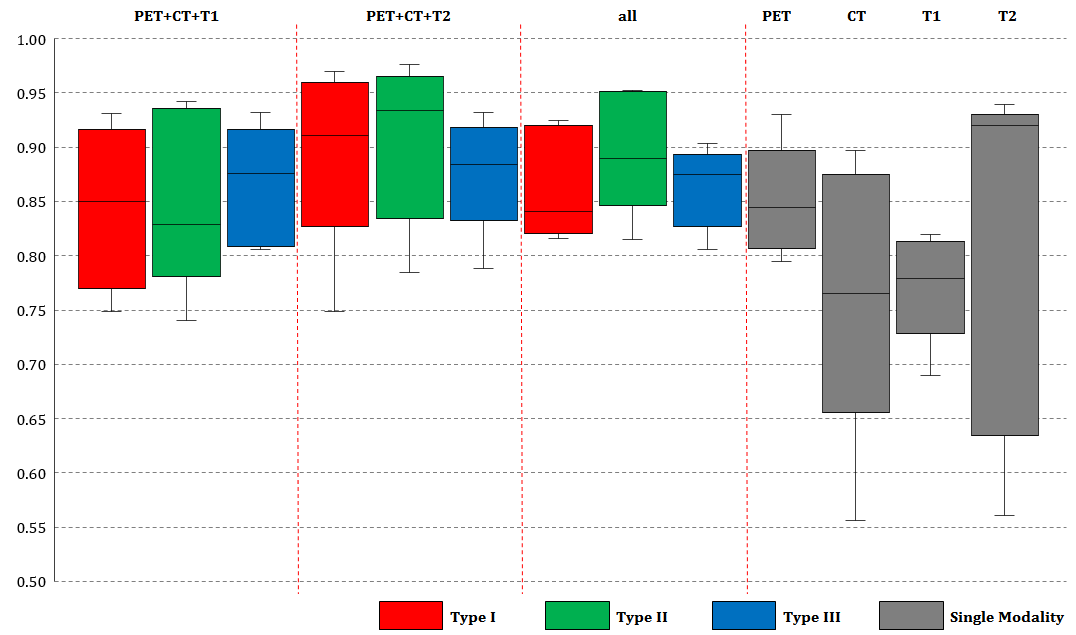}
		\caption{Box chart for the statistics (median, first/third quartile and the min/max) of the pixel-wise prediction accuracies. Each box corresponds to one specific type of network trained and tested on one specific combination of modalities. For example, the first box from the left shows the statistics of prediction accuracy of Type-I fusion network trained and tested on images from PET, CT and T1-weighted MR imaging modalities. }
	\end{centering}
\end{figure}

From the statistics, it can be found that multi-modality network show superior performance in tumor segmentation than single-modality networks. Among all the single-modality networks, the one trained and tested on the T2-weighted MR has the best performance, since the annotation is mainly based on T2 images (better identification of soft tissue), as illustrated in Fig. 1. It is observed that the performance of PET-based network is the worst on average while PET is designed to detect the tumor presence. This is mainly caused by the necrosis in the center of large tumor which barely show uptake in FDG-PET images.

Further, although the annotation is mainly determined on T2-weighted images, the fusion networks trained and tested on the combination of PET, CT and T1 (without T2) can achieve better accuracy comparing with single-modality network based on T2. Such result shows that while a single modality might be inaccurate and/or insufficient to characterize the tumor region by itself, the fusion network can automatically take advantage of the combined information. A sample result of such case is shown in Fig. 5, where the multi-modality fusion network (Fig. 5b) can obtain the better result comparing with T2-based single modality network (Fig. 5c). A closer examination of the single modality networks based on PET, CT and T1 shows that neither of these three modalities can lead to a good prediction: PET (Fig. 5d) suffers from the necrosis in the center issue as discuss above, while a large region of false positive is presented in CT, T1 and T2 results (Fig. 5c, e, and f). 

\begin{figure}[H]
	\begin{centering}
		\includegraphics[width=1.0\textwidth]{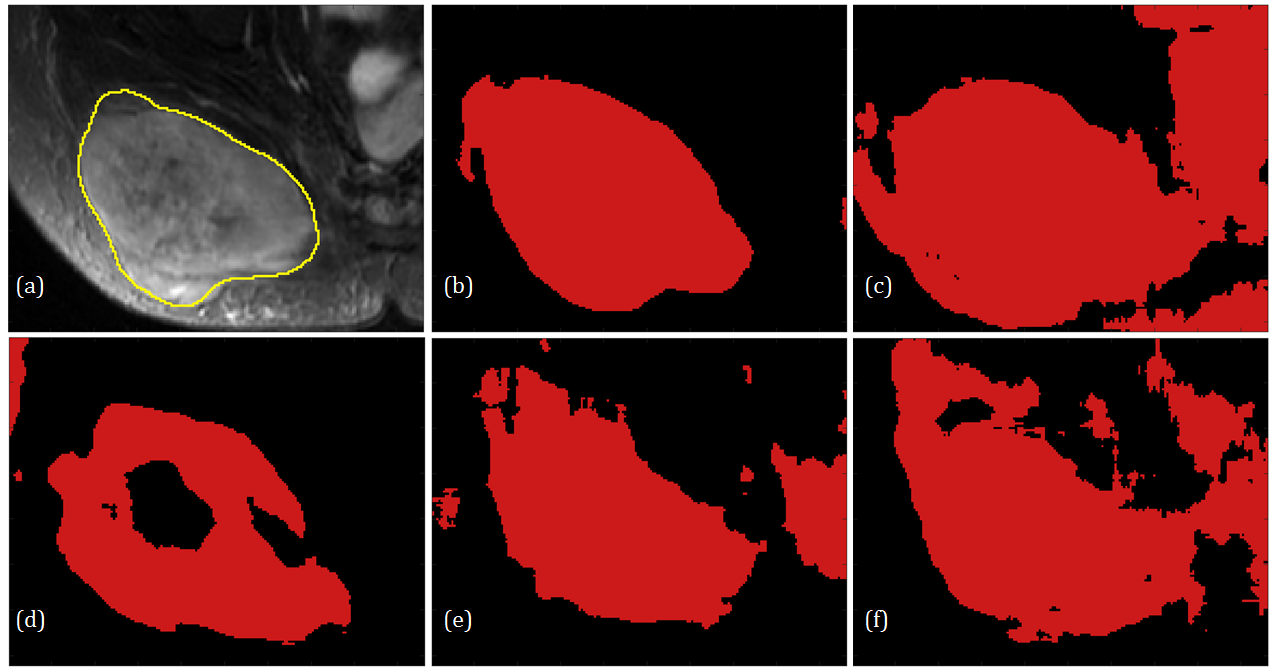}
		\caption{(a) Ground truth shown as yellow contour line overlaid on the T2 image. (b) Result from Type-II fusion network based on PET+CT+T1. (c) Result from single-modality network based on T2. (d-f) Results from single-modality network based on PET, CT and T1, respectively.}
	\end{centering}
\end{figure}  

\subsection{Discussion of fusion schemes}

Among all the schemes, Type-I and Type-II fusion networks have shown similarly good performance as in Fig. 4. Yet the computational cost for training Type-I fusion network is much less than Type-II, which is due to their differences in the network structure where Type-II fusion network will need to learn much more number of features than type-I fusion network. Empirical evaluation shows that the training time for Type-I network is only one-third to Type-II network when analyzing three image modalities and around one-fourth when analyzing four modalities, which is a huge advantage. More generally, learning a single fused feature set from multi-modal data is easier and can be done in an integrated framework. In addition, fusing at decision level performs the worst among all fusion schemes, as majority vote usually fails to capture the intrinsic relationship between image modalities. However, it is advantageous when the single-modal classification is readily available as no extra computation is needed to perform the fusion.

From a learning representation perspective, Type-I fusion network learns the lowest-level features from different modalities separately at the first layer, but immediately fuses them based on convolutional kernel to learn the higher-level features at the second layer. On the contrary, Type-II fusion network learns one set of low-to-high features for each modality, then uses fully connected network to combine the feature outputs. One implication for such differences is that Type-II fusion network can learn at least one set of useful features given that at least one modality is informative, while features learned by Type-I fusion network can be more severely disrupted. We have observed various cases from the experimental results that bad training samples can result in the unstable performance and decreased robustness for Type-I fusion network, as indicated by the large variance shown in Fig. 4. Specifically, the presence of a bad image modality which is either not related with the tumor region and/or has lowered signal-to-noise ratio (SNR) can disrupt the multi-modal learning. One example is visualized in Fig. 6. In this example, the performance of Type-I fusion network (Fig. 6e) is much worse than Type-II fusion network (Fig. 6f). As single-modality network trained on the PET image almost produces opposite results to the ground truth (Fig. 6d), we hypothesize that the PET modality is the reason for the worse performance of type-I fusion network, but not affecting type-II fusion network as much.

\begin{figure}[H]
	\begin{centering}
		\includegraphics[width=1.0\textwidth]{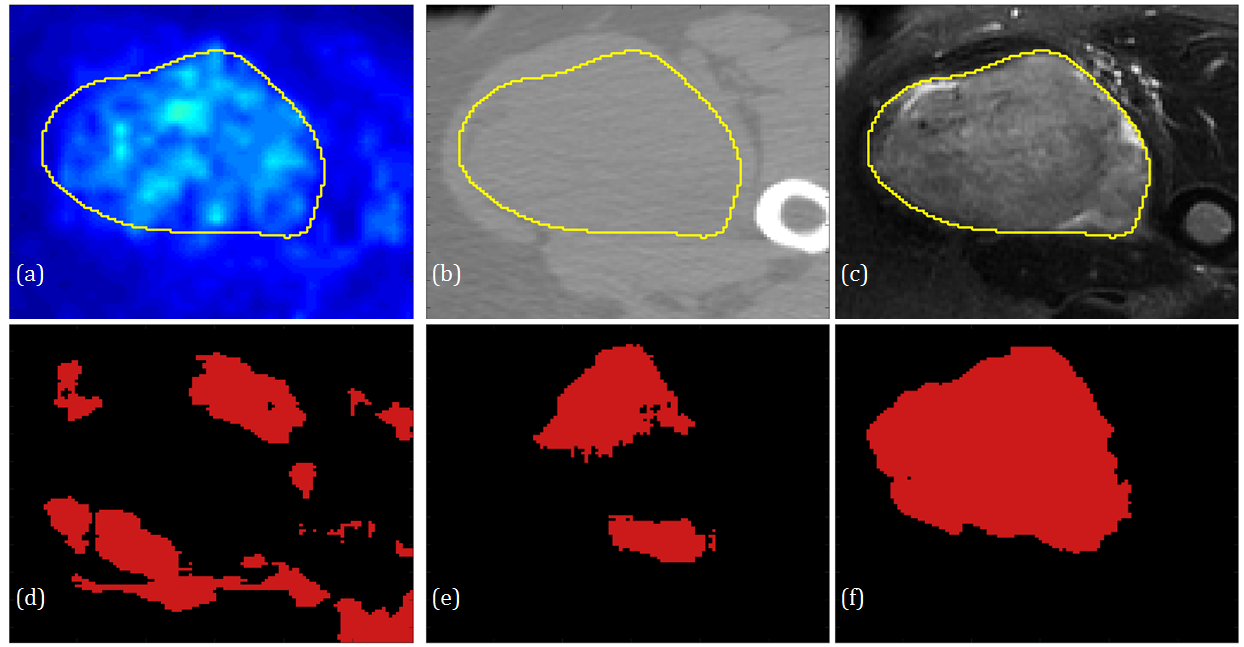}
		\caption{(a) Ground truth shown as yellow contour line overlaid on the input PET image. (b) Input CT image. (c) Input T2 image. (d) Results from single-modality network based on PET. (e) Result from Type-I fusion network based on PET+CT+T2. (f) Result from Type-II fusion network based on PET+CT+T2.}
	\end{centering}
\end{figure}

\section{Conclusion and Discussion}
In this work, we propose a generalized framework of image fusion for supervised learning in biomedical image analysis which consists of three fusion schemes, and implement the fusion schemes based on deep convolutional neural network. The fusion networks as well as their single-modality counterparts are then tested on the TCIA Soft-tissue-Sarcoma dataset for the purpose of computer-aided detection. In addition to the supreme performance obtained by the fusion networks, suggesting its usefulness for other types of multi-modal medical image analysis, we also discuss the possible rationale behind the performance differences among the three fusion schemes and provide insights into the image fusion problem in general. While our conclusions are only preliminary and based on empirical observation, we envision that the error at data and/or label within image modalities will have different algorithmic impact on the learning process. Ultimately, such intrinsic difference will cause the performance of different fusion schemes depending on the applications and image qualities variations across modalities.

\section{Acknowledgement}
The first author of this work is supported by the China Scholarship Council. This work is also supported by the MGH \& BWH Center for Clinical Data Science (CCDS), and we especially would like to acknowledge the support of NVIDIA Corporation with the donation of the DGX1 to the CCDS which is used for this research. We gratefully acknowledge the help from Prof. Junwei Han of Northwestern Polytechnical University, China and his students in the algorithm development and implementation of this work. 

\pagebreak

\bibliographystyle{unsrt}
\bibliography{manuscript_MMCNN}

\begin{thebibliography}{10}

\bibitem{RN908}
B.~H. Menze, A.~Jakab, S.~Bauer, J.~Kalpathy-Cramer, K.~Farahani, J.~Kirby,
  Y.~Burren, N.~Porz, J.~Slotboom, R.~Wiest, L.~Lanczi, E.~Gerstner, M.~A.
  Weber, T.~Arbel, B.~B. Avants, N.~Ayache, P.~Buendia, D.~L. Collins,
  N.~Cordier, J.~J. Corso, A.~Criminisi, T.~Das, H.~Delingette, Demiralp Ç,
  C.~R. Durst, M.~Dojat, S.~Doyle, J.~Festa, F.~Forbes, E.~Geremia, B.~Glocker,
  P.~Golland, X.~Guo, A.~Hamamci, K.~M. Iftekharuddin, R.~Jena, N.~M. John,
  E.~Konukoglu, D.~Lashkari, J.~A. Mariz, R.~Meier, S.~Pereira, D.~Precup,
  S.~J. Price, T.~R. Raviv, S.~M.~S. Reza, M.~Ryan, D.~Sarikaya, L.~Schwartz,
  H.~C. Shin, J.~Shotton, C.~A. Silva, N.~Sousa, N.~K. Subbanna, G.~Szekely,
  T.~J. Taylor, O.~M. Thomas, N.~J. Tustison, G.~Unal, F.~Vasseur,
  M.~Wintermark, D.~H. Ye, L.~Zhao, B.~Zhao, D.~Zikic, M.~Prastawa, M.~Reyes,
  and K.~Van Leemput.
\newblock The multimodal brain tumor image segmentation benchmark (brats).
\newblock {\em IEEE Transactions on Medical Imaging}, 34(10):1993--2024, 2015.

\bibitem{RN916}
Michael Czisch, Thomas~C. Wetter, Christian Kaufmann, Thomas Pollmächer,
  Florian Holsboer, and Dorothee~P. Auer.
\newblock Altered processing of acoustic stimuli during sleep: Reduced auditory
  activation and visual deactivation detected by a combined fmri/eeg study.
\newblock {\em NeuroImage}, 16(1):251--258, 2002.

\bibitem{RN907}
Alex~Pappachen James and Belur~V. Dasarathy.
\newblock Medical image fusion: A survey of the state of the art.
\newblock {\em Information Fusion}, 19(Supplement C):4--19, 2014.

\bibitem{RN901}
M.~Vallières, C.~R. Freeman, S.~R. Skamene, and I.~El Naqa.
\newblock A radiomics model from joint fdg-pet and mri texture features for the
  prediction of lung metastases in soft-tissue sarcomas of the extremities.
\newblock {\em Physics in Medicine \& Biology}, 60(14):5471, 2015.

\bibitem{RN763}
Nicolle~M. Correa, Tom Eichele, Tülay Adal, Yi-Ou Li, and Vince~D. Calhoun.
\newblock Multi-set canonical correlation analysis for the fusion of concurrent
  single trial erp and functional mri.
\newblock {\em NeuroImage}, 50(4):1438--1445, 2010.

\bibitem{RN906}
Marco Lorenzi, Ivor~J. Simpson, Alex~F. Mendelson, Sjoerd~B. Vos, M.~Jorge
  Cardoso, Marc Modat, Jonathan~M. Schott, and Sebastien Ourselin.
\newblock Multimodal image analysis in alzheimer\'s disease via statistical
  modelling of non-local intensity correlations.
\newblock {\em Scientific Report}, 6:22161, 2016.

\bibitem{RN909}
Xinzheng Xu, Dong Shan, Guanying Wang, and Xiangying Jiang.
\newblock Multimodal medical image fusion using pcnn optimized by the qpso
  algorithm.
\newblock {\em Appl. Soft Comput.}, 46(C):588--595, 2016.

\bibitem{RN904}
G.~Bhatnagar, Q.~M.~J. Wu, and Z.~Liu.
\newblock Directive contrast based multimodal medical image fusion in nsct
  domain.
\newblock {\em IEEE Transactions on Multimedia}, 15(5):1014--1024, 2013.

\bibitem{RN905}
Rajiv Singh and Ashish Khare.
\newblock Fusion of multimodal medical images using daubechies complex wavelet
  transform: A multiresolution approach.
\newblock {\em Information Fusion}, 19(Supplement C):49--60, 2014.

\bibitem{RN910}
Y.~Yang.
\newblock Multimodal medical image fusion through a new dwt based technique.
\newblock In {\em 2010 4th International Conference on Bioinformatics and
  Biomedical Engineering}, pages 1--4, 2010.

\bibitem{RN896}
Xiaofeng Zhu, Heung-Il Suk, Seong-Whan Lee, Dinggang Shen, and Initiative the
  Alzheimer\'s Disease~Neuroimaging.
\newblock Subspace regularized sparse multi-task learning for multi-class
  neurodegenerative disease identification.
\newblock {\em IEEE transactions on bio-medical engineering}, 63(3):607--618,
  2016.

\bibitem{RN839}
N.~Liu, J.~Han, T.~Liu, and X.~Li.
\newblock Learning to predict eye fixations via multiresolution convolutional
  neural networks.
\newblock {\em IEEE Transactions on Neural Networks and Learning Systems},
  PP(99):1--13, 2017.

\bibitem{RN912}
Stefan Klein, Uulke~A. van~der Heide, Irene~M. Lips, Marco van Vulpen, Marius
  Staring, and Josien P.~W. Pluim.
\newblock Automatic segmentation of the prostate in 3d mr images by atlas
  matching using localized mutual information.
\newblock {\em Medical Physics}, 35(4):1407--1417, 2008.

\bibitem{RN913}
H.~Cai, R.~Verma, Y.~Ou, S.~k. Lee, E.~R. Melhem, and C.~Davatzikos.
\newblock Probabilistic segmentation of brain tumors based on multi-modality
  magnetic resonance images.
\newblock In {\em 2007 4th IEEE International Symposium on Biomedical Imaging:
  From Nano to Macro}, pages 600--603, 2007.

\bibitem{RN902}
Kenneth Clark, Bruce Vendt, Kirk Smith, John Freymann, Justin Kirby, Paul
  Koppel, Stephen Moore, Stanley Phillips, David Maffitt, Michael Pringle,
  Lawrence Tarbox, and Fred Prior.
\newblock The cancer imaging archive (tcia): Maintaining and operating a public
  information repository.
\newblock {\em Journal of Digital Imaging}, 26(6):1045--1057, 2013.

\end{thebibliography}

\end{document}